\documentclass[11pt]{article}
\usepackage{txfonts}
\usepackage{graphicx}

\title{Harnessing Natural Fluctuations: Analogue Computer for Efficient Socially Maximal Decision Making}
\author{Song-Ju Kim${}^{1\dag}$, Makoto Naruse${}^{2}$ and Masashi Aono${}^{3,4}$\\
\ \\
\noindent
\hspace*{-5mm}
${}^{1}$ WPI Center for Materials Nanoarchitectonics (MANA),\\
\hspace*{-5mm}National Institute for Materials Science (NIMS),\\
\hspace*{-5mm}1--1 Namiki, Tsukuba, Ibaraki 305--0044, Japan\\
\ \\
\noindent
\hspace*{-5mm}${}^{2}$ Photonic Network Research Institute,\\
\hspace*{-5mm}National Institute of Information and Communications Technology\\
\hspace*{-5mm}4--2--1 Nukui-kita, Koganei, Tokyo 184--8795, Japan\\
\ \\
\noindent
\hspace*{-5mm}${}^{3}$ Earth-Life Science Institute, Tokyo Institute of Technology\\
\hspace*{-5mm}2--12--1 Ookayama, Meguro-ku, Tokyo 152--8550, Japan\\
\hspace*{-5mm}${}^{4}$ PRESTO, Japan Science and Technology Agency\\ 
\hspace*{-5mm}4--1--8 Honcho, Kawaguchi-shi, Saitama 332--0012, Japan\\
\ \\
\noindent
\hspace*{-5mm}${}^{\dag}$KIM.Songju@nims.go.jp
}

\begin{document}

\maketitle

\newpage

\begin{abstract}
Each individual handles many tasks of finding the most profitable option from a set of options that stochastically provide rewards. 
Our society comprises a collection of such individuals, and the society is expected to maximise the total rewards, while the individuals compete for common rewards. 
Such collective decision making is formulated as the `competitive multi-armed bandit problem (CBP)', requiring a huge computational cost.
Herein, we demonstrate a prototype of an analog computer that efficiently solves CBPs by exploiting the physical dynamics of numerous fluids in coupled cylinders.
This device enables the maximisation of the total rewards for the society without paying the conventionally required computational cost; this is because the fluids estimate the reward probabilities of the options for the exploitation of past knowledge and generate random fluctuations for the exploration of new knowledge.
Our results suggest that to optimise the social rewards, the utilisation of fluid-derived natural fluctuations is more advantageous than applying artificial external fluctuations.
Our analog computing scheme is expected to trigger further studies for harnessing the huge computational power of natural phenomena for resolving a wide variety of complex problems in modern information society.
\end{abstract}

\section*{Introduction}

The benefits to an organization (the whole) and those to its constituent members (parts) sometimes conflict. For example, let us consider a situation wherein traffic congestion is caused by a driver making a selfish decision to pursue his/her individual benefit to quickly arrive at a destination. In a situation wherein a car bound from south to north approaches an intersection where preceding vehicles are stalled while the signal is about to turn in red, the driver must refrain from selfishly deciding to enter the intersection. Otherwise, the car would obstruct other vehicles' paths in the west and east directions, stalled in the intersection after the signal turned red. Thus, the whole's benefit can be spoiled by that of a part. 

The conflict between the whole's benefit and that of the parts frequently arises in a wide variety of situations in modern society. Confrontations between communities and wars between nations can be seen as caused by collisions of global and local interests.
%Is it extremely visionary to think that human beings who have agreed to develop the `equipment which derives an overall optimisation solution' and to follow it can face a new age wherein it can minimise barren confrontations?
In realistic political judgment, many of these collisions are modelled using a game-theoretic approach by appropriately setting up a payoff matrix ~\cite {mesquita}. In mobile communication, the channel assignment problem in cognitive radio communication can also be represented as a particular class of payoff matrix. Herein, we consider the competitive bandit problem (CBP), which a problem of maximising total rewards through collective decision making and requires a huge computational cost for an increase in problem size. We demonstrate a method for exploiting the computational power of the physical dynamics of numerous fluids in coupled cylinders to efficiently solve the problem.

Consider two slot machines. Both machines have individual reward probabilities $P_A$ and $P_B$. At each trial, a player selects one of the machines and obtains some reward, a coin for example, with the corresponding probability. The player wants to maximise the total reward sum obtained after a particular number of selections. However, it is assumed that the player does not know these probabilities. How can the player gain maximal rewards? The multi-armed bandit problem (BP) involves determining the optimal strategy for selecting the machine which yields maximum rewards by referring to past experiences.

\begin{figure}[ht]
\vspace{30mm}
\centering
\hspace*{-60mm}
\includegraphics[height=60mm, width=50mm, bb= 0 0 500 600]{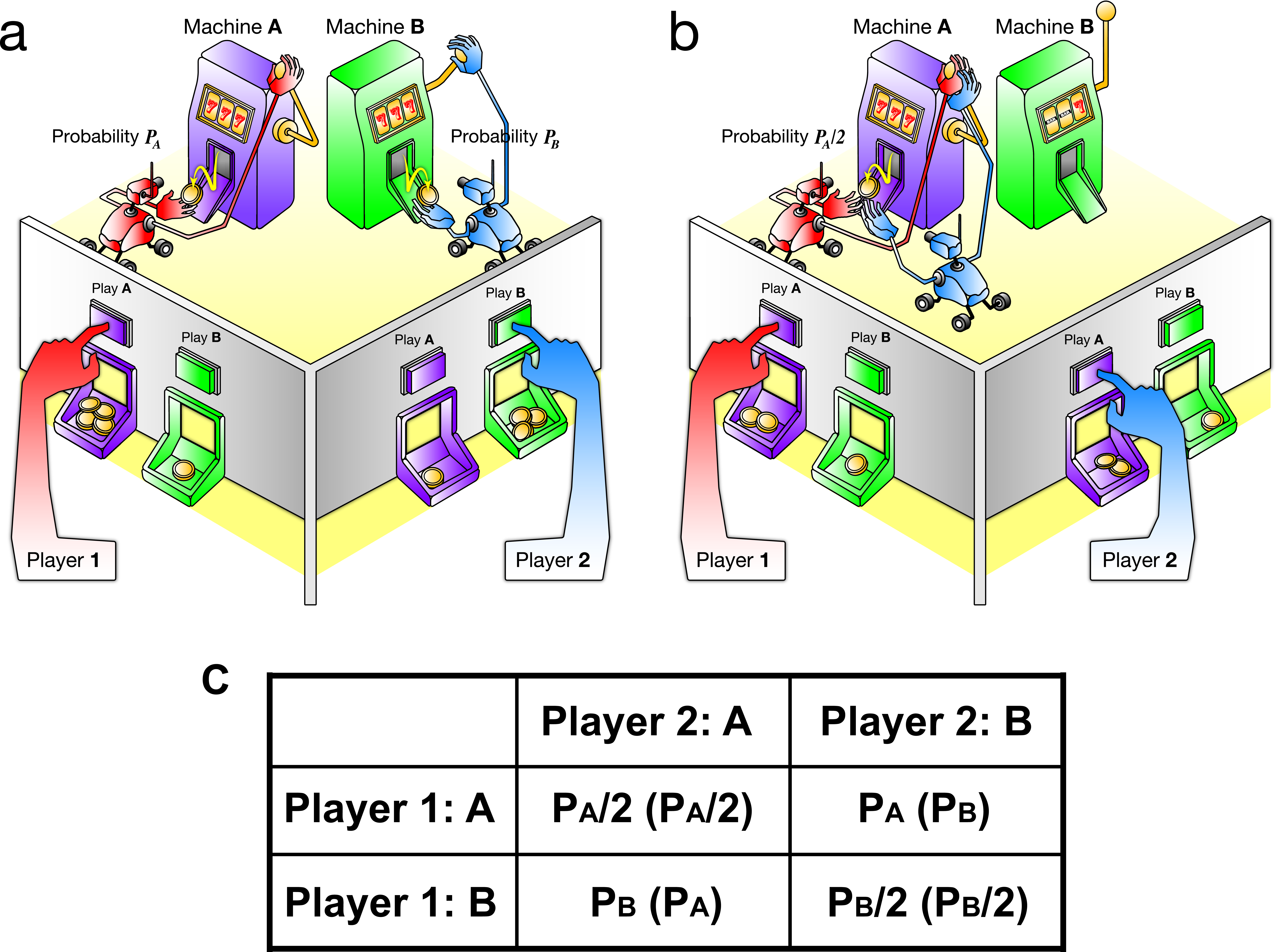}
\centering
\caption{Competitive Bandit Problem (CBP). (a) segregation state. (b) collision state. (c) Payoff matrix for player 1 (player 2).}
\label{fig:CBP}
\end{figure}
For simplicity, we consider here the minimum CBP, i.e. two players (1 and 2) and two machines ($A$ and $B$), as shown in Fig.~\ref{fig:CBP}. It is supposed that a player playing a machine can obtain some reward, a coin for example, with the probability $P_i$. Figure~\ref{fig:CBP}(c) shows the payoff matrix for players 1 and 2.

If a collision occurs, i.e. two players select the same machine, the reward is evenly split between those players. We seek an algorithm that can obtain the maximum total rewards (scores) of all players. To acquire the maximum total rewards, the algorithm must contain a mechanism that can avoid the `Nash equilibrium' states, which are the natural consequence for a group of independent selfish players, and can determine the `social maximum~\cite{game}' states which can obtain maximum total rewards. In our previous studies~\cite{kim1,kim2,kimNOLTA,kim11,aono,kimCMA,kimP}, we showed that our proposed algorithm called `Tug-Of-War (TOW) dynamics' is more efficient than other well-known algorithms such as the modified $\epsilon$-greedy and softmax algorithms, and is comparable to the `upper confidence bound1-tuned (UCB1T) algorithm', which is known as the best among parameter-free algorithms~\cite{auer}. Moreover, TOW dynamics effectively adapts to a changing environment wherein the reward probabilities dynamically switch.
Algorithms for solving CBP are applicable to various fields such as Monte Carlo tree search, which is used in algorithms for the `game of GO'~\cite{uct,mogo}, cognitive radio~\cite{cog,cog2}, and web advertising~\cite{web}.

Herein, by applying TOW dynamics that exploit the volume conservation law, we propose a physical device that efficiently computes the optimal machine assignments of all players in a centralised control. The proposed device consists of two kinds of fluids in cylinders: one representing `decision making by a player' and the other representing the `interaction between players (collision avoider)'. 
%Since this device is based on simple dynamics of physical objects called a `tug-of-war (TOW) dynamics' and satisfies the volume conservation law, it can be physically implemented simply by using two kinds of incompressible fluids in two or more cylinders. 
We call the physical device the `TOW bombe' owing to its similarity to the `Turing bombe' invented by Alan Turing, the analog electric circuit used by the British army for decoding the German army's `enigma code' of the during World War II ~\cite{bom}. The assignment problem for $M$ players and $N$ machines can be automatically solved simply by repeatedly operating (up-and-down operation of the fluid interface in a cylinder) $M$ times at every iteration in the TOW bombe without calculating the evaluation values of $O(N^M)$. This suggests that an analog computer is more advantageous than a digital computer, if we appropriately use the natural phenomena. Although the problems considered here are not really nondeterministic-polynomial-time (NP) problems, we can show advantages of natural fluctuations generated in the device and suggest a possibility to extend the device to apply to NP problems. 
%The extension of the device and solvable problems is also discussed in discussion and SI. 
Using the TOW bombe, we can automatically achieve the social maximum assignments by entrusting the huge amount of computations for evaluation values to the physical processes of fluids.

%\section{TOW Dynamics}

%In previous studies~\cite{kim11,kimCMA,kimP}, we proposed the Tug-of-War (TOW) dynamics.
\begin{figure}[h]
\vspace{50mm}
\centering
\hspace*{-40mm}
\includegraphics[height=60mm, width=40mm, bb= 0 0 400 600]{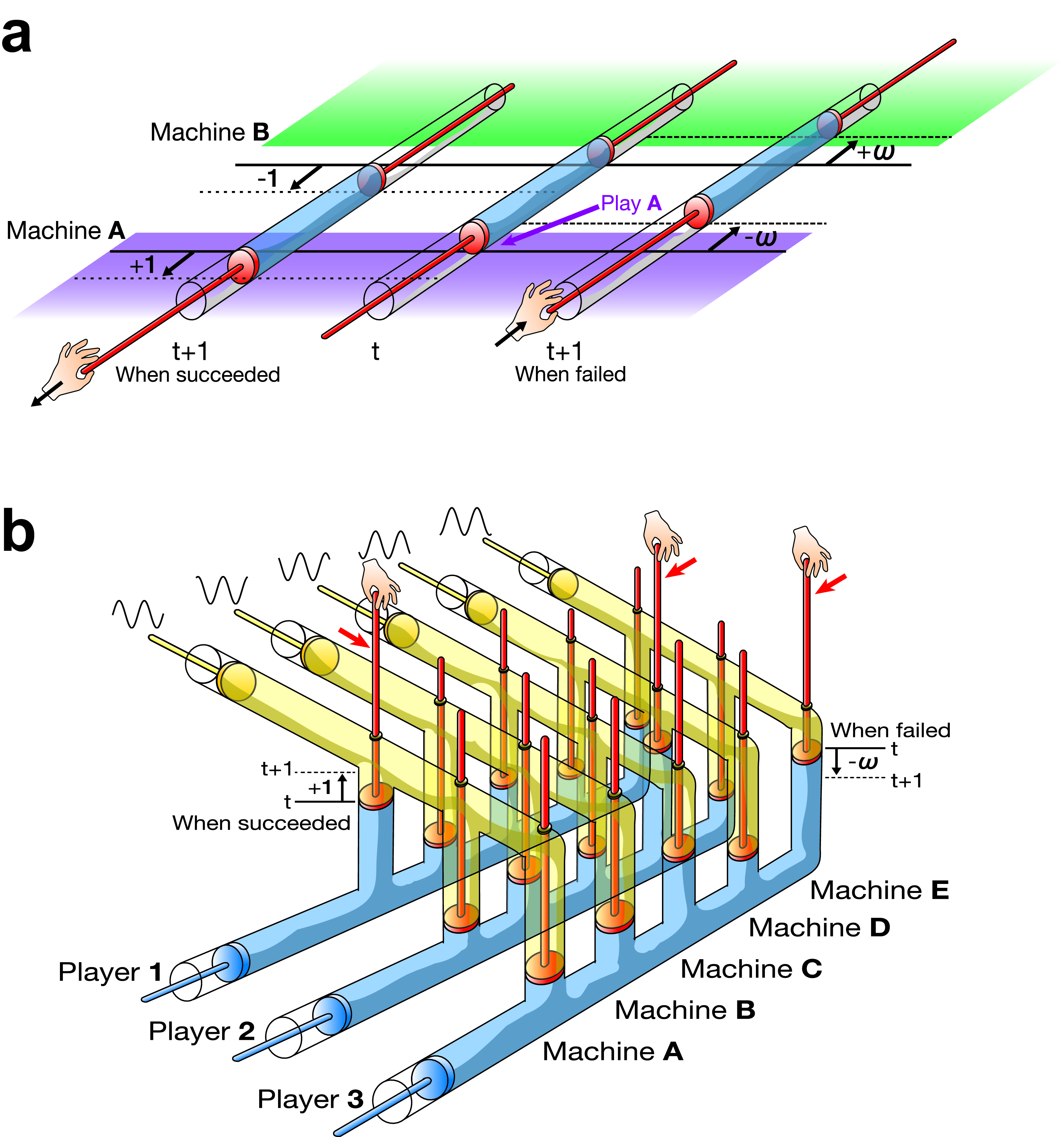}
\caption{(a) TOW dynamics. (b) The TOW bombe for three players and five channels.}
\label{fig:bombe}
\end{figure}
Consider an incompressible fluid in a cylinder, as shown in Fig.~\ref{fig:bombe}(a). Here, $X_k$ corresponds to the displacement of terminal $k$ from an initial position, where $k\in \{A,B\}$. If $X_k$ is greater than $0$, we consider that the liquid selects machine $k$.
%In TOW dynamics, the MBP is represented in its inverse form, as we introduce `punishment' instead of `reward'.
%That is, when machine $k$ is played, the player is `punished' with a probability $1 - P_k$.

We used the following estimate $Q_k$ ($k\in \{A,B\}$):
\begin{equation}
Q_k(t)  =   N_k(t) - (1 + \omega) L_k(t). \label{qq1}
\end{equation}
Here, $N_k$ is the number of playing machine $k$ until time $t$ and $L_k$ is the number of non-rewarded (i.e. failed) events in $k$ until time $t$, where $\omega$ is a weighting parameter (see Method).

The displacement $X_A$ ($= - X_B$) is determined by the following difference equation:
\begin{eqnarray}
X_A(t) & = & Q_A(t) - Q_B(t) + \delta .\label{eq:os}
\end{eqnarray}
Here, $\delta(t)$ is an arbitrary fluctuation to which the liquid is subjected. Consequently the TOW dynamics evolve according to a particularly simple rule: in addition to the fluctuation, if machine $k$ is played at each time $t$, $+1$ and $-\omega$ are added to $X_k(t-1)$ when rewarded and non-rewarded, respectively (Fig.~\ref{fig:bombe}(a)). The authors have shown that this simple dynamics gains more rewards (coins or packet transmissions in cognitive radio) than those obtained by other popular algorithms for solving the BP~\cite{kim1,kim2,kimNOLTA,kim11,aono,kimCMA,kimP}.

Many algorithms for the BP estimate the reward probability of each machine. In most cases, this `estimate' is updated only when the corresponding machine is selected. In contrast, TOW dynamics uses a unique learning method which is equivalent to that updating both estimates simultaneously owing to the volume conservation law~\cite{kimNOLTA,kim11}. TOW dynamics can imitate the system that determines its next moves at time $t+1$ in referring to the estimate of each machine, even if it was not selected at time $t$, as if the two machines were simultaneously selected at time $t$. This unique feature is one of the sources of the TOW's high performance~\cite{kimP}.
We call this the `TOW principle'.
This principle is also applicable to a more general BP (see Method).

\section{Results}

%\section{The TOW Bombe}

The TOW bombe for three players ($1, 2$ and $3$) and five machines ($A, B, C, D$ and $E$) is illustrated in Figure~\ref {fig:bombe}(b). Two kinds of incompressible fluids (blue and yellow) fill coupled cylinders. The blue (bottom) fluid handles a player's decisions made, while the yellow (upper) one handles interaction among players. Machine selection of each player at each iteration is determined by the height of a red adjuster (a fluid interface level), and the highest machine is chosen. When the movements of blue and yellow adjusters stabilise to reach equilibrium, the TOW principle in the blue fluid holds for each player. In other words, when one interface rises, the other four interfaces fall, resulting in efficient machine selections. Simultaneously, the action-reaction law holds for the yellow fluid (i.e. if the interface level of player 1 rises, the interface levels of players 2 and 3 fall), contributing collision avoidance, and the TOW bombe can search for an overall optimisation solution accurately and quickly. In normal use, however, blue and yellow adjusters must have fixed positions not to move. 

The dynamics of the TOW bombe are expressed as follows: 
\begin{eqnarray}
Q_{(i,k)}(t)  & = &  \Delta Q_{(i,k)}(t) + Q_{(i,k)}(t-1) \nonumber \\
 & & - \frac{1}{M-1}\sum_{j \neq i} \Delta Q_{(j,k)}(t), \\
X_{(i,k)}(t) & = & Q_{(i,k)}(t) - \frac{1}{N-1} \sum_{l \neq k} Q_{(i,l)}(t). 
\end{eqnarray} 
Here, $X_{(i, k)}(t) $ is the height of the interface of player $i$ and machine $k$ at iteration step $t$. If machine $k$ is chosen for player $i$ at time $t$, $\Delta Q_{(i, k)}(t)$ is $+1$ or $-\omega $ according to the result (rewarded or not). Otherwise, it is $0$. 

In addition to the above-mentioned dynamics, some fluctuations or external oscillations are added to $X_{(i, k)}$. These added fluctuations or oscillations are sensitive to the TOW bombe's performance, because fluctuations represent exploration patterns in the early stage.

Thus, the TOW bombe operates only by adding an operation which raises or lowers the interface level ($+1$ or $-\omega$) according to the result (success or failure of coin gain) for each player (total $M$ times) at each time. After these operations, the interface levels move according to the volume conservation law, calculating the next selection for each player. In each player's selection, an efficient search is achieved as a result of the TOW principle, which can obtain a solution accurately and quickly for trial-and-error tasks. Moreover, through the interaction among players via yellow fluid, the Nash equilibrium can be avoided, thereby achieving the social maximum~\cite{game}. 

%\section{Results}

To show that the TOW bombe avoids the Nash equilibrium and regularly achieves an overall optimisation, we consider a case wherein ($P_A$, $P_B$, $P_C$, $P_D$, $P_E$) $=$ ($0.03$, $0.05$, $0.1$, $0.2$, $0.9$) as a typical example. For simplicity, part of the payoff tensor that has $125$ (=$5^3$) elements is described as follows; only matrix elements for which each player does not choose low-ranking $A$ and $B$ are shown (Table~\ref {table:2}, \ref {table:3} and \ref {table:4}). For each matrix element, the reward probabilities are given in the order of players 1, 2 and 3. 
\begin{table}[ht]
\caption{Payoff matrix of the case where ($P_C$, $P_D$, $P_E$)$=$($0.1$, $0.2$, $0.9$), player 3 chooses $C$}
\label{table:2}
\begin{center}
\begin{tabular}{|c|c|c|c|} \hline \hline
          & player 2: C        & player 2: D        & player 2: E \\ \hline
player 1: C & $1/30$, $1/30$, $1/30$ & $0.05$, $0.2$, $0.05$ & $0.05$, $0.9$, $0.05$ \\ \hline
player 1: D & $0.2$, $0.05$, $0.05$ & $0.1$, $0.1$, $0.1$ & $0.2$, $0.9$, $0.1$ {\bf SM} \\\hline
player 1: E & $0.9$, $0.05$, $0.05$ & $0.9$, $0.2$, $0.1$ {\bf SM} & $0.45$, $0.45$, $0.1$ \\\hline
\end{tabular}
\end{center}
\end{table}
\begin{table}[h]
\caption{Payoff matrix of the case where ($P_C$, $P_D$, $P_E$)$=$($0.1$, $0.2$, $0.9$), player 3 chooses $D$}
\label{table:3}
\begin{center}
\begin{tabular}{|c|c|c|c|} \hline \hline
          & player 2: C        & player 2: D        & player 2: E \\ \hline
player 1: C & $0.05$, $0.05$, $0.2$ & $0.1$, $0.1$, $0.1$ & $0.1$, $0.9$, $0.2$ {\bf SM} \\ \hline
player 1: D & $0.1$, $0.1$, $0.1$ & $2/30$, $2/30$, $2/30$ & $0.1$, $0.9$, $0.1$ \\\hline
player 1: E & $0.9$, $0.1$, $0.2$ {\bf SM} & $0.9$, $0.1$, $0.1$ & $0.45$, $0.45$, $0.2$ \\\hline
\end{tabular}
\end{center}
\end{table}
\begin{table}[h]
\caption{Payoff matrix of the case where ($P_C$, $P_D$, $P_E$)$=$($0.1$, $0.2$, $0.9$), player 3 chooses $E$}
\label{table:4}
\begin{center}
\begin{tabular}{|c|c|c|c|} \hline \hline
          & player 2: C        & player 2: D        & player 2: E \\ \hline
player 1: C & $0.05$, $0.05$, $0.9$ & $0.1$, $0.2$, $0.9$ {\bf SM} & $0.1$, $0.45$, $0.45$ \\ \hline
player 1: D & $0.2$, $0.1$, $0.9$ {\bf SM} & $0.1$, $0.1$, $0.9$ & $0.2$, $0.45$, $0.45$ \\\hline
player 1: E & $0.45$, $0.1$, $0.45$ & $0.45$, $0.2$, $0.45$ & $0.3$, $0.3$, $0.3$ {\bf NE} \\\hline
\end{tabular}
\end{center}
\end{table}

Social maximum ({\bf SM}) is a state in which the maximum amount of total reward is obtained by all the players. In this problem, the social maximum corresponds to a segregation state in which the players choose the top three distinct machines ($C, D, E$), respectively; there are six segregation states indicated by {\bf SM} in the Tables. In contrast, the Nash equilibrium ({\bf NE}) is a state in which all the players choose machine $E$ independent of others' decisions; machine $E$ gives the reward with the highest probability, when each player behaves selfishly.

\begin{figure}[h]
\vspace{60mm}
\centering
\hspace*{-120mm}
\includegraphics[height=20mm, width=30mm, bb= 0 0 300 200]{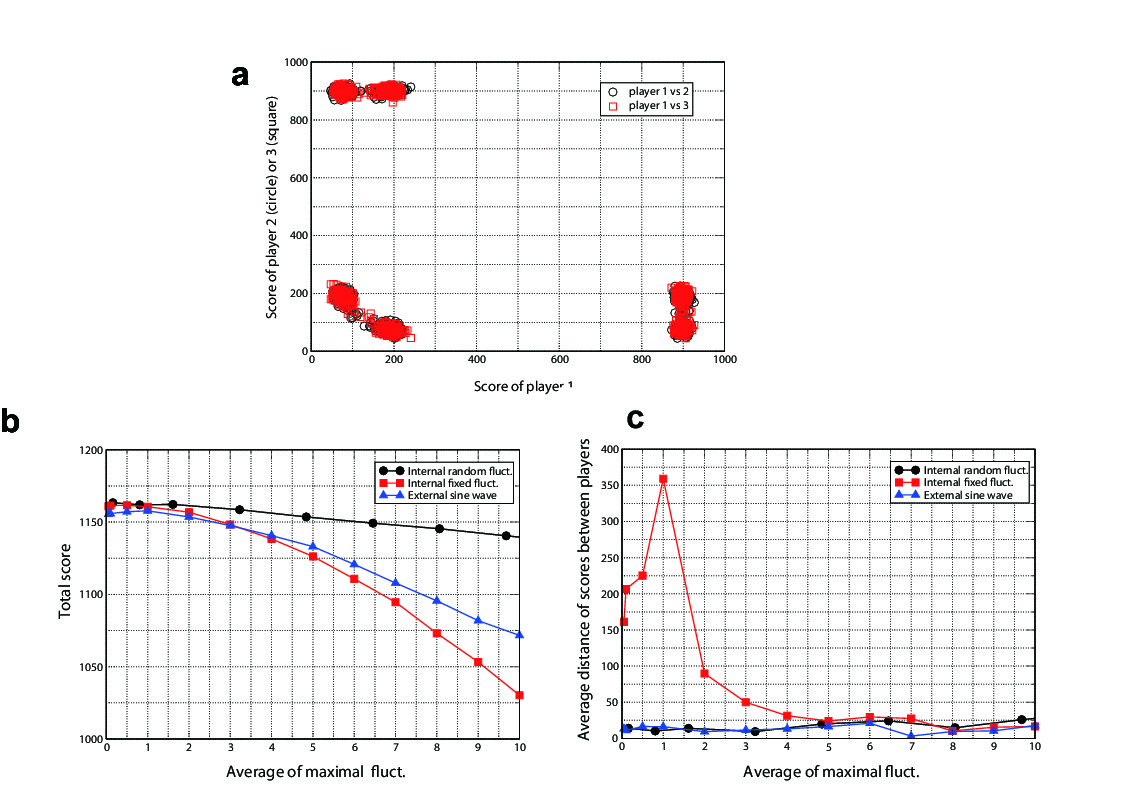}
\centering
\vspace{2mm}
\caption{(a) TOW bombe scores in the case wherein ($P_A$, $P_B$, $P_C$, $P_D$, $P_E$) $=$ ($0.03$, $0.05$, $0.1$, $0.2$, $0.9$).
(b) Sample averages of total TOW bombe scores in the case wherein ($P_A$, $P_B$, $P_C$, $P_D$, $P_E$) $=$ ($0.03$, $0.05$, $0.1$, $0.2$, $0.9$). (c) Sample averages of mean distance between players' scores in the case wherein ($P_A$, $P_B$, $P_C$, $P_D$, $P_E$) $=$ ($0.03$, $0.05$, $0.1$, $0.2$, $0.9$).}
\label{fig:dots}
\end{figure}

The performance of the TOW bombe was evaluated using a score: the number of rewards (coins) a player obtained in his/her $1,000$ plays. In cognitive radio communication, the score corresponds to the number of packets that have successfully transmitted~\cite{cog,cog2}. Figure~\ref {fig:dots}(a) shows the TOW bombe scores in the typical example wherein ($P_A$, $P_B$, $P_C$, $P_D$, $P_E$) $=$ ($0.03$, $0.05$, $0.1$, $0.2$, $0.9$). Since $1,000$ samples were used, there are $1,000$ circles. Each circle indicates the score obtained by player $i$ (horizontal axis) and player $j$ (vertical axis) for one sample. 
There are six clusters in Figure~\ref {fig:dots}(a) corresponding to the two-dimensional projections of the six segregation states, implying the overall optimisation. The social maximum points are given as follows: (the score of player $1$, the score of player $2$, the score of player $3$) $=$ ($100$, $200$, $900$), ($100$, $900$, $200$), ($200$, $100$, $900$), ($200$, $900$, $100$), ($900$, $100$, $200$) and ($900$, $200$, $100$). The TOW bombe did not reach the Nash equilibrium state ($300$, $300$, $300$).
%Here, we set weighting parameter $\omega$ at $0.08$ (Eq. (\ref {w0m1}) and (\ref {w0m2}) 
%were calculated as $\gamma^{\prime}$$=$$P_B$$+$$P_C$). 

In our simulations, we used `adaptive' weighting parameter $\omega$, meaning that the parameter is estimated by using its own variables (see Method). Owing to this estimation cost, clusters of circles are not located exactly at the social maximum points. If we set weighting parameter $\omega$ at $0.08$, which are calculated as $\gamma^{\prime}$$=$$P_B$$+$$P_C$ (see Method), those clusters are located exactly on the social maximum points (see Figures in \cite{kimASTE}).

Figure~\ref{fig:dots}(b) shows TOW bombe performance, sample averages of the total scores of all players up to $1,000$ plays, for three different type of fluctuation, respectively. The black, red and blue lines denote the cases of internal random fluctuations, internal fixed fluctuations and external oscillations, respectively (see Method). The horizontal axis denotes the sample averages of maximum fluctuation. In the maximal case, the average total score has gained nearly $1,200$ ($=$$100$$+$$200$$+$$900$), which is the value of the social maximum, although there are some gaps resulting from estimation costs.

Figure~\ref{fig:dots}(c) also shows TOW bombe fairness, sample averages of the mean distance between players' scores, for three different types of fluctuation, respectively. We can confirm lower fairness in the cases of internal fixed fluctuations (red line). Artificially created fluctuations, such as internal fixed fluctuations, often show lower fairness because of the existence of biases (lack of uniformity or randomness) in fluctuations. Although the external oscillations (sine waves) have higher fairness (blue line), controlling the blue and yellow adjusters appropriately is difficult. Moreover, the performances of these two types of fluctuation rapidly decrease as the magnitude of fluctuations increases, as shown in Fig.~\ref{fig:dots}(b).  

We can conclude that only the internal random fluctuations, which are supposed to be generated automatically in the real TOW bombe, exhibit higher performance and fairness.
This conclusion is consistent even in cases where we set weighting parameter $\omega$ at $0.08$.
This indicates the construction of a novel analog computing scheme which exploits nature's power in terms of automatic generation of random fluctuations, simultaneous computations using a conservation law and intrinsic efficiency. 

\section{Discussion}
%\section{Extension of the TOW bombe for NP problems}

How can we harness nature's power for computations such as automatic generation of random fluctuations, simultaneous computations using a conservation law and intrinsic efficiency as well as the feasibility of massive computations?

Alan Turing mathematically clarified a concept of `computation' by proposing his Turing machine, the most simple model of computation~\cite{Turing1,Turing2}.
%This `simplicity' is for human beings but not nature.
A Turing machine consists of a sequence of steps which can read and write a single symbol on tape. These `discrete' and `sequential' steps are `simple' for a human to understand. Moreover, he found a `universal Turing machine' that can simulate all other computations. Owing to this machine, algorithms can be studied on their own, without regard to the systems that are implementing them ~\cite{moore}. Human beings no longer need to be concerned about underlying mechanisms. In other words, software can be abstracted away from hardware. This property has brought substantial development in digital computers. Simultaneously, however, these algorithms have lost links to natural phenomena implementing them. He had exchanged natural affinity for artificial convenience. 

Digital computers created a `monster' called `exponential explosion', wherein computational cost grows exponentially as a function of problem size (NP problems). In our daily lives, we often encounter this type of problem, such as scheduling, satisfiability (SAT) and resource allocation problems. For a digital computer, such problems become intractable as the problem size grows. In contrast, nature always `computes' infinitely many computations at every moment~\cite{Feynman}. However, we do not know how to extract and harness this power of nature.

Herein, we demonstrate that an analog decision-making device, called the TOW bombe, can be implemented physically by using two kinds of incompressible fluid in coupled cylinders and can efficiently achieve overall optimisation in the machine assignment problem in CBP by exploiting nature's power, including automatic generation of random fluctuations, simultaneous computations using a conservation law and intrinsic efficiency. The randomness of fluctuations generated automatically in the real TOW bombe might not be high, but there are ways to enhance randomness. For example, turbulence occurs if we move an adjuster rapidly in an up-and-down operation.  
%infinitesimally small
% adiabatic transition

The TOW bombe enables us to solve the assignment problem for $M$ players and $N$ machines by repeating $M$ up-and-down operations of fluid interface levels in cylinders at each iteration; it does not require calculation for as many evaluation values as are required when using a conventional digital computer, because it entrusts the huge amount of computation to the physical processes of fluids. This suggests that there are advantages to analog computation even in today's digital age.

%Actually, the CBP is not an NP problem.
Although the payoff tensor has $N^M$ elements, the TOW bombe need not hold $N^M$ evaluation values. If we ignore the diagonal elements, $N$ evaluation values are sufficient for each player's estimation of which machine is the best. Therefore, using the TOW bombe, the CBP is reducible to an $O(NM)$ problem when implementing a collision-avoiding mechanism handled by yellow fluid, although, in a strict sense, the computational cost must include the cost for providing random fluctuations generated by the fluids' physical dynamics. In Fig.~\ref{fig:dots}(b), we showed the results of only three types of fluctuation. TOW bombe performance with internal $M$-random fluctuations (see Method) was the same as that of the internal random fluctuations, although computations for generating the former type of fluctuation require a cost that exponentially grows as $O(N^M)$. This is because the exponential type of fluctuation is not effective for $O(NM)$ problems. Various random seed patterns do not affect enhancing the performance of $O(NM)$ problems because of the reducibility of CBP to three independent BPs.

However, this is not the cases if we focus on more complex problems, such as the `Extended Prisoner's Dilemma Game' (see Supplementary Information); we must prepare more than $N M$ evaluation values, because a player's reward is drastically changed according to the selections of other players in this problem. There are some cases that can be approximately solved by the TOW bombe even in this type of complex problem. In these cases, the exponential type of fluctuation can enhance the performance slightly. This fact may suggest that we find the first toehold to harnessing nature's power which is the feasibility of massive computations.

Unfortunately, it is difficult to solve this type of complex problem using the TOW bombe in general. To solve more complex problems, we must also extend the TOW bombe. We have some ideas regarding TOW bombe extension using some fluid compressibility, local inflow and outflow, a reservoir for blue or yellow fluid, a time order of fluctuations and quantum effects such as non-locality and entanglement. The TOW bombe can also be implemented on the basis of quantum physics. In fact, the authors have exploited optical energy transfer dynamics between quantum dots and single photons to design decision-making devices~\cite{qdm,qdm2,qdm3}. Our method might be applicable to a class of problems derived from CBP and broader varieties of game payoff tensors, implying that wider applications can be expected. We will report these observations and results elsewhere in the future.

\section*{Methods}

\subsection*{The weighting parameter $\omega$}

TOW dynamics involves the parameter $\omega$ which is sensitive to its performance. From analytical calculations, it is known that the following $\omega_0$ is sub-optimal in the BP (see Supplementary Information or \cite{kimP}),
\begin{eqnarray}
\omega_0 & = & \frac{\gamma}{2 - \gamma} ,\\
\gamma & = & P_A + P_B .
\end{eqnarray}
Here, it is assumed that $P_A$ is the largest reward probability and $P_B$ is the second largest.

In the CBP cases ($M$-player and $N$-machine), the following $\omega_0$ is sub-optimal,
\begin{eqnarray}
\omega_0  & = & \frac{\gamma^{\prime}}{2 - \gamma^{\prime}} , \\
\gamma^{\prime} & = & P_{(M)} + P_{(M+1)}
\end{eqnarray} 
Here, $P_{(M)}$ is the top $M$th reward probability.

Players must estimate $\omega_0$ using its variables, because information regarding reward probabilities is not given to players. We call this an `adaptive' weighting parameter. There are many estimate methods, such as Bayesian inference, but we simply use `direct substitution' herein. Direct substitution uses $R_j(t) / N_j(t)$ for $P_j$, where $R_j(t)$ is the number of reward gains from machine $j$ through time $t$ and $N_j(t)$ is the number of plays of machine $j$ through time $t$. 

\subsection*{TOW dynamics for general BP}

In this paper, we use TOW dynamics only for the Bernoulli type of BP in which the reward $r$ is $1$ or $0$. 
Another type of TOW dynamics can also be constructed for general BP in which the reward $r$ is a real value from an interval $[0, R]$.
Here, $R$ is arbitrary positive value, and the reward $r$ is selected according to given probability distribution whose mean and variance are $\mu$ and $\sigma^2$, respectively.

In this case, the following estimate $Q_k$ ($k\in \{A,B\}$) is used insted of eq.(\ref{qq1}):
\begin{equation}
Q_k(t)  =   \Sigma^t_{j=1} r_k(j) - \gamma^{*} N_k(t). \label{gbp}
\end{equation}
Here, $N_k$ is the number of playing machine $k$ until time $t$ and $r_k(j)$ is the reward in $k$ at  time $j$, where $\gamma^{*}$ is the following parameter:
\begin{equation}
\gamma^{*}  =   \frac{\mu_A + \mu_B}{2}.
\end{equation}
If machine $k$ is played at each time $t$, the reward $r_k(t)$ and $-\gamma^{*}$ are added to $X_k(t-1)$.

\subsection*{Generating methods of fluctuation}

\subsubsection*{1. Internal fixed fluctuations}

First, we define fixed moves $O_{k^{\prime}}$ ($k^{\prime}={0, \cdots, 4}$), as follows, 
\begin{equation}
 \{ O_{0}, O_{1}, O_{2}, O_{3}, O_{4} \} = \{ 0, A, 0, -A, 0 \}  
\end{equation}
Here, $A$ is an amplitude parameter.
Note that $\sum^{4}_{k^{\prime}=0}$ $O_{k^{\prime}}$ $=$ $0$.

To use the above move $O_{k}$ recursively, we introduce a new variable $num$ ($num={0, \cdots, 4}$), as follows, 
\begin{equation}
num = \{ t + (k-2) \} \hspace{2mm} mod \hspace{2mm} 5 
\end{equation}
Here, $t$ is a time.
For each machine $k$ ($k={1, \cdots, 5}$), we use the following set of fluctuations, respectively, 
%for(k=1;k<=5;k++){
\begin{eqnarray}
 osc_{(1,k)}(t) & = & O_{0}, \\
 osc_{(2,k)}(t) & = & O_{3}, \\
 osc_{(3,k)}(t) & = & O_{1}.
\end{eqnarray}
If $num$ $=$ $0$. 

\begin{eqnarray}
 osc_{(1,k)}(t) & = & O_{1}, \\
 osc_{(2,k)}(t) & = & O_{4}, \\
 osc_{(3,k)}(t) & = & O_{3}.
\end{eqnarray}
If $num$ $=$ $1$.
 
\begin{eqnarray}
 osc_{(1,k)}(t) & = & O_{2}, \\
 osc_{(2,k)}(t) & = & O_{0}, \\
 osc_{(3,k)}(t) & = & O_{4}.
\end{eqnarray}
If $num$ $=$ $2$. 

\begin{eqnarray}
 osc_{(1,k)}(t) & = & O_{3}, \\
 osc_{(2,k)}(t) & = & O_{1}, \\
 osc_{(3,k)}(t) & = & O_{2}.
\end{eqnarray}
If $num$ $=$ $3$.
 
\begin{eqnarray}
 osc_{(1,k)}(t) & = & O_{4}, \\
 osc_{(2,k)}(t) & = & O_{2}, \\
 osc_{(3,k)}(t) & = & O_{0}.
\end{eqnarray}
If $num$ $=$ $4$. 

It always holds that $\sum^{3}_{i=1}$ $osc_{(i,k)}(t)$ $=$ $0$ and $\sum^{5}_{k=1}$ $osc_{(i,k)}(t)$ $=$ $0$. These conditions mean that added fluctuations to $X_{(i, k)}$ can be cancelled in total. In other words, the total volume of blue or yellow fluid does not change. As a result, we create artificial `internal' fluctuations.

\subsubsection*{2. Internal random fluctuations}

First, a matrix sheet of random fluctuations ($Sheet_{(i,k)}$) is prepared.
Here, $i$ $=$ ${1, \cdots, 3}$ and $k$ $=$ ${1, \cdots, 5}$.
\begin{enumerate}
\item
$r$ is a random value from $[0, 1]$.
We call this `seed'.
\item
There are $N M$ ($ = 15$) possibilities for a seed position. Choose the seed position ($i_0$, $k_0$) randomly from $i_0$ $=$ ${1, \cdots, 3}$ and $k_0$ $=$ ${1, \cdots, 5}$ and place the seed $r$ at the point, 
\begin{equation}
Sheet_{(i_0,k_0)} = r .
\end{equation}
\item
All elements of the $k_0$th column other than ($i_o$, $k_0$) are substituted with $-0.5*r$.
\item
All elements of the $i_0$-th row other than ($i_o$, $k_0$) are substituted with $-0.25*r$.
\item
All remaining elements are substituted with $r/8.0$.
\item
The matrix sheet is accumulated in a summation matrix $Sum_{(i,k)}$.   
\item
Repeat from two to six for $D$ times. Here, $D$ is a parameter.
\end{enumerate}            
We used the following set of fluctuations, 
\begin{equation}
osc_{(i,k)}(t) = A/D * Sum_{(i,k)} .
\end{equation}
Here, $A$ is an amplitude parameter.

It always holds that $\sum^{3}_{i=1}$ $osc_{(i,k)}(t)$ $=$ $0$ and $\sum^{5}_{k=1}$ $osc_{(i,k)}(t)$ $=$ $0$, as well as the internal fixed fluctuations. The total volume of blue or yellow fluid does not change. As a result, we create `internal' random fluctuations naturally. At every time step, this procedure costs $O(N \cdot M)$ computations with a digital computer.

\subsubsection*{3. Internal $M$-random fluctuations (exponential)}

First, a matrix sheet of random fluctuations ($Sheet_{(i,k)}$) is prepared.
Here, $i$ $=$ ${1, \cdots, 3}$ and $k$ $=$ ${1, \cdots, 5}$.
\begin{enumerate}
\item
For each player $i$, independent random value $r_i$ is generated from $[0, 1]$. We call these `seeds'.
\item
There are $N^M$ ($ = 125$) possibilities for a seed position pattern. For each player $i$, choose the seed position ($i$, $k_0(i)$) randomly from $k_0(i)$ $=$ ${1, \cdots, 5}$ and place the seed $r_i$ at the point, 
\begin{equation}
Sheet_{(i,k_0(i))} = r_i .
\end{equation}
However, we choose $k_0(i)$s to be distinct.
Therefore, there are really $N (N-1) (N-2)$ ($=60$) possibilities.
\item
For each $i$, all elements of the $k_0(i)$-th column other than ($i$, $k_0(i)$) are substituted with $-0.5*r_i$.
\item
All remaining elements of the $1$th row are substituted with $-0.50*(r_1 - 0.50*r_2 - 0.50*r_3)$.
\item
All remaining elements of the $2$ th row are substituted with $-0.50*(r_2 - 0.50*r_1 - 0.50*r_3)$.
\item
All remaining elements of the $3$ th row are substituted with $-0.50*(r_3 - 0.50*r_1 - 0.50*r_2)$.
\item
The matrix sheet is accumulated in a summation matrix $Sum_{(i,k)}$.   
\item
Repeat from two to seven for $D$ times. Here, $D$ is a parameter.
\end{enumerate}            
We used the following set of fluctuations, 
\begin{equation}
osc_{(i,k)} = A/D * Sum_{(i,k)} .
\end{equation}
Here, $A$ is an amplitude parameter.

It always holds that $\sum^{3}_{i=1}$ $osc_{(i,k)}(t)$ $=$ $0$ and $\sum^{5}_{k=1}$ $osc_{(i,k)}(t)$ $=$ $0$ as well as the internal fixed or random fluctuations. The total volume of blue or yellow fluid does not change. As a result, we create `internal' $M$-random fluctuations naturally. At every time step, this procedure costs exponential computations of $O(N^M)$ with a digital computer.

\subsubsection*{4. External oscillations}

Herein, we used completely synchronised oscillations $osc_{(i, k)}(t) $ added to every player's $X_{(i, k)}$,   
\begin{equation}
osc_{(i,k)}(t) = A \hspace{1mm} sin(2 \pi t / 5 + 2 \pi (k-1)/5) .
\end{equation}
Here, $i = 1$, $\cdots$, $3$ and $k = 1$, $\cdots$, $5$. 
$A$ is an amplitude parameter.
These oscillations are externally provided by appropriately controlling the blue and yellow adjusters. 

\section*{Acknowledgement} 

%This study partially includes the work authors S. -J. K. and M. A. performed at the time of enrolment in the RIKEN Advanced Science Institute. We thank Prof. Masahiko Hara and Dr. Etsushi Nameda for valuable discussions and advice.
%Moreover, we are grateful to Prof. Hirokazu Hori at Yamanashi University for useful argument about the theory of the TOW bombe and its quantum extension. 
This work was supported in part by the Sekisui Chemical Grant Program for `Research on Manufacturing Based on Innovations Inspired by Nature'.
We are grateful to Prof. Hirokazu Hori at Yamanashi University for useful argument about the theory of the TOW bombe and its quantum extension. 

\section*{Contributions}
S.-J.K. and M.A. designed the research. S.-J.K. designed and simulated the TOW Bombe. S.-J.K., M.A. and M.N. analysed the data. M.N advised quantum extension of the model. S.-J.K. and M.A wrote the manuscript. All authors reviewed the manuscript.

\section*{Competing financial interests}

The authors declare that there is no conflicting financial interest regarding the publication of this paper. 
%All authors have filed a patent application on the technology described herein.

%\newpage
%
%\begin{figure}[h]
%\centering
%\includegraphics[height=80mm]{EDFIG1.eps}
%\caption{(a) Scores of the TOW bombe with $\omega=0.08$ in the case where ($P_A$, $P_B$, $P_C$, $P_D$, $P_E$) $=$ ($0.03$, $0.05$, $0.1$, $0.2$, $0.9$). (b) Sample averages of the scores of the TOW bombe with $\omega=0.08$ in the case where ($P_A$, $P_B$, $P_C$, $P_D$, $P_E$) $=$ ($0.03$, $0.05$, $0.1$, $0.2$, $0.9$). (c) Scores of the TOW bombe with $\omega=0.33$ where ($P_A$, $P_B$, $P_C$, $P_D$, $P_E$) $=$ ($0.1$, $0.2$, $0.3$, $0.4$, $0.5$). (d) Sample averages of the scores of the TOW bombe with $\omega=0.33$ in the case where ($P_A$, $P_B$, $P_C$, $P_D$, $P_E$) $=$ ($0.1$, $0.2$, $0.3$, $0.4$, $0.5$).}
%\label{fig:aste}
%\end{figure}
%
%\newpage
%
%\begin{figure}[h]
%\centering
%\includegraphics[height=80mm]{EDFIG2.eps}
%\vspace{2mm}
%\caption{Sample averages of total scores of the TOW bombe in the Extended Prisoner's Dilemma Game}
%\label{fig:totalN}
%\end{figure}

\newpage

\begin{center}
{\Huge {\bf Supplementary Information}}
\end{center}
\vspace{5mm}

\section*{Efficient Decision-Making by Physical Objects}

Elements of computing devices are subject to physical laws that work as `constraints'. These constraints always have negative effects on computing ability.
%For example, in the CMOS structure, not-a-simple circuit is required even for a simple NAND or NOR operation due to deficient freedom caused by the constraints.
For example, in a complementary metal oxide semiconductor (CMOS) structure, considerably complicated circuits are required even for simple logical operations such as NAND and NOR because physical constraints can violate logically correct behaviour in simple structures.
 
However, herein, we show the opposite fact that such constraints can also have positive effects. That is, computational efficiency can be generated from the movements of physical objects which are subjected to the volume conservation law.
This `tug-of-war (TOW) principle' is addressed as efficiency in `trial-and-error'.
 
Consider two slot machines. Both machines have individual reward probabilities $P_A$ and $P_B$. At each trial, a player selects one of the machines and obtains some reward, a coin for example, with the corresponding probability. The player wants to maximize the total reward sum obtained after a particular number of selections. However, it is supposed that the player does not know these probabilities. The multi-armed bandit problem (BP) involves determining the optimal strategy for selecting the machine which yields maximum rewards by referring to past experiences.

The BP was originally described by Robbins~\cite{robbins}, although the essential problem was studied earlier by Thompson ~\cite{thompson}. The optimal strategy is known only for a limited class of problems wherein the reward distributions are assumed to be `known' to the players~\cite{gittins1,gittins2}. Even in these problems, computing the Gittins index in practice is not tractable for many problems. Agrawal and Auer et al. proposed decision-making algorithms that could express the index as a simple function of the total reward obtained from a machine~\cite{agra,auer}. Especially, the `upper confidence bound 1 (UCB1) algorithm' proposed by Auer is used worldwide for many applications.

Kim et al. proposed a `decision-making dynamics' called TOW; it was inspired by the true slime mold {\it Physarum}~\cite{kim1,kim2,kim3,kim4,kim5,kim6}, which maintains a constant intracellular resource volume while collecting environmental information by concurrently expanding and shrinking its branches. 
The conservation law entails a `non-local correlation' among the branches, that is, the volume increment in one branch is immediately compensated for by volume decrement(s) in the other branch(es). This non-local correlation was shown to be useful for decision making.
%Thus, TOW is a dynamical system that describes spatiotemporal dynamics of a physical object (i.e. an amoeboid organism). TOW dynamics connected `natural phenomena' to `decision-making' for the first time. This approach enables us to achieve an `efficient decision maker', an object that can efficiently make a decision. In fact, Kim et al. proposed a decision maker using optical energy transfer between quantum dots (QDs) that can solve the BP efficiently~\cite{QDM,QDM2,QDM3}.

In this paper, we propose `the TOW principle', which explains why computational efficiency can be generated from the movements of physical objects that are subjected to the volume conservation law.

\subsection*{Solvability: random walk approach}

Let us consider a one-dimensional random walk, where the distance of right flight when `a coin' is dispensed is $\alpha$ and the distance of left flight when `no coin' is dispensed is $\beta$. We assume that $P_A$ (probability of right flight in random walk $A$) $>$ $P_B$ (probability of right flight in random walk $B$) for simplicity. After time step $t$, the displacement $R_k(t)$ ($k\in \{A,B\}$) can be described by
\begin{eqnarray}
R_k(t) & = & \alpha (N_k - L_k) - \beta \hspace{1mm}L_k  \nonumber \\
       & = & \alpha N_k - (\alpha + \beta)\hspace{1mm}L_k . \label{eq:ran}
\end{eqnarray}
Here, $N_k$ is the number of playing machine $k$ until time $t$ and $L_k$ is the number of non-rewarded (i.e. left flight) events in $k$ until time $t$.
The expected value of $R_k$ can be obtain from the following equation,
\begin{equation}
E( R_k(t) ) = \{\alpha P_A - \beta (1 - P_B)\}\hspace{1mm} N_k .
\end{equation}

In the overlapping area between two distributions of $R_A$ and $R_B$, we cannot estimate correctly which is the greater. The overlapping area must decrease as $N_k$ increases to avoid incorrect judgements. This requirement can be expressed in the following forms:
\begin{eqnarray}
  \alpha P_A - \beta (1 - P_B)     & > & 0 ,  \\
  \alpha P_B - \beta (1 - P_A)     & < & 0 . \label{eq:cond1}
\end{eqnarray}
These forms can be transformed to the form  
\begin{equation}
P_B  < \frac{\beta}{\alpha + \beta}   < P_A .  \label{eq:cond1}
\end{equation}
In other words, the parameter $\alpha$ and $\beta$ must satisfy the above conditions for the random walk to represent correctly the larger judgement. We can easily confirm that the following form satisfies these conditions
\begin{equation}
\frac{\beta}{\alpha + \beta}   = \frac{P_A + P_B}{2} .\label{eq:fixed}
\end{equation}

On the other hand, we use the following learning rule in our TOW dynamics, 
\begin{equation}
Q_k(t)  =   N_k(t) - (1 + \omega) L_k(t). \label{qq}
\end{equation}
Here, $\omega$ is a weighting parameter.
From $R_k(t)/\alpha $ $=$ $Q_k(t)$, we obtain
\begin{equation}
\omega = \frac{\beta}{\alpha} . \label{eq:walpha}
\end{equation}
From Eq.(\ref{eq:fixed}) and (\ref{eq:walpha}), we can obtain
\begin{eqnarray}
\omega & = & \frac{\gamma}{2 - \gamma} ,\\
\gamma & = & P_A + P_B .
\end{eqnarray}
Therefore, we can conclude that the algorithm using the learning rule $Q_k$ with the parameter $\omega$ can accurately solve the BP. Here, we use $\omega_0$ for the above $\omega$. Detailed analytical calculations are presented in \cite{kimP}.

\subsection*{The TOW principle}

In many popular algorithms, such as the $\epsilon$-greedy algorithm, an estimate for reward probability is updated only in a selected arm. In contrast, we consider the case wherein the sum of the reward probabilities $\gamma$ $=$ $P_A$ $+$ $P_B$ is given. Then, we can update both estimates simultaneously as follows,

\vspace{1mm}
\begin{center}
\begin{tabular}{cccc}
$A$: & \LARGE{$\frac{N_A - L_A}{N_A}$} & $B$: & $\gamma \hspace{1mm} -$ \LARGE{$\frac{N_A - L_A}{N_A}$}, \\ \\
$A$: & $\gamma \hspace{1mm} -$ \LARGE{$\frac{N_B - L_B}{N_B}$} & $B$: & \LARGE{$\frac{N_B - L_B}{N_B}$}. 
\end{tabular}
\end{center}
\vspace{1mm}

\noindent
Here, the top and bottom rows give estimates based on the information that machines $A$ and $B$ were selected $N_A$ and $N_B$ times, respectively. Note that we can also update the estimate of the machine that was not played, owing to information $\gamma$.
 
From the above estimates, each expected reward $Q^{\prime}_k$ ($k\in \{A,B\}$) is given as follows,
\begin{eqnarray}
Q^{\prime}_A & = & N_A \hspace{1mm} \frac{N_A - L_A}{N_A} + N_B \hspace{1mm} \bigl( \gamma \hspace{1mm} - \frac{N_B - L_B}{N_B} \bigr) \nonumber \\
 & = & N_A - L_A + (\gamma - 1) \hspace{1mm} N_B + L_B, \label{eq:qAp}\\
Q^{\prime}_B & = & N_A\hspace{1mm} \bigl( \gamma \hspace{1mm} - \frac{N_A - L_A}{N_A} \bigr) + N_B\hspace{1mm} \frac{N_B - L_B}{N_B} \nonumber \\
& = & N_B - L_B + (\gamma - 1) \hspace{1mm} N_A + L_A. \label{eq:qBp}
\end{eqnarray}
These expected rewards $Q^{\prime}_j$s are not the same as the TOW dynamics learning rules, $Q_j$s (Eq.(\ref{qq})). However, what we use substantially in the TOW is the difference,
\begin{equation}
Q_A - Q_B  =  (N_A - N_B) - (1 + \omega)\hspace{1mm}(L_A - L_B)\label{eq:dq}.
\end{equation}
When we transform the expected rewards $Q^{\prime}_j$s into 
\begin{eqnarray}
Q^{\prime \prime}_A & = & Q^{\prime}_A / (2 - \gamma), \label{eq:qApp}\\
Q^{\prime \prime}_B & = & Q^{\prime}_B / (2 - \gamma), \label{eq:qBpp}
\end{eqnarray}
we can obtain the difference,
\begin{equation}
Q^{\prime \prime}_A - Q^{\prime \prime}_B  =  (N_A - N_B) - \frac{2}{2-\gamma} \hspace{1mm} (L_A - L_B). \label{eq:dqpp}
\end{equation}
Comparing the coefficient of Eqs.(\ref{eq:dq}) and (\ref{eq:dqpp}), those two differences are always equal when $\omega=\omega_0$ satisfies,
\begin{equation}
\omega_0  =  \frac{\gamma}{2-\gamma}. \label{eq:w0}
\end{equation} 
Eventually, we can obtain the nearly optimal weight parameter $\omega_0$ in terms of $\gamma$.

This derivation means that TOW dynamics has a learning rule equivalent to that of the system that can simultaneously update both estimates. TOW can imitate the system that determines its next moves at time $t+1$ in referring to the estimate of each arm even if it was not selected at time $t$, as if the two arms were simultaneously selected at time $t$. This unique feature in the learning rule, derived from the fact that the sum of reward probabilities is given in advance, is one of the sources of TOW's high performance.

Performing Monte Carlo simulations, it was confirmed that the performance of TOW dynamics with $\omega_0$ is comparable to its best performance, i.e. the TOW with $\omega_{opt}$. To derive the $\omega_{opt}$ accurately, we need to consider the fluctuations~\cite{kim4}.

The essence described here can be extended to general $K$ machines cases. If you want to separate distributions of the top $m$th and $(m+1)$th machine in the previous subsection, all you need do is use the following parameter $\omega_0$.
\begin{eqnarray}
\omega_0  & = & \frac{\gamma^{\prime}}{2 - \gamma^{\prime}} , \\
\gamma^{\prime} & = & P_{(m)} + P_{(m+1)}
\end{eqnarray} 
Here, $P_{(m)}$ denotes the top $m$th reward probability.

\subsection*{Discussion}

Performances of algorithms that can solve the BP are mostly evaluated using the `$regret$' defined as follows, which quantifies the accumulated losses of rewards,
\begin{equation}
regret = (P_A - P_B) E(N_B).
\end{equation}
Here, $E(N_B)$ denotes the expected value of $N_B$.

It is known that optimal algorithms for the BP, defined by Auer et al., have a regret proportional to $\log(N)$~\cite{auer}. The regret has no finite upper bound as $N$ increases because it continues to require playing the lower-reward machine to ensure that the probability of incorrect judgment goes to zero. Interestingly, we analytically demonstrated in our previous work that our TOW dynamics has a constant $regret$ ~\cite{kimP}. A constant regret means that the probability of incorrect judgment remains non-zero in TOW dynamics, though this probability is nearly equal to zero. However, it would appear that the reward probabilities change frequently in actual decision-making situations and their long-term behaviour is not crucial for many practical purposes. For this reason, TOW dynamics would be more suited to real-world applications.

Herein, we propose `the TOW principle' which explains why computational efficiency can be generated from the movements of physical objects due to the volume conservation law. In ordinal decision-making algorithms, the parameter of `exploration time' is optimized for the BP. That parameter ordinarily represents $|$ $P_A$ $-$ $P_B$ $|$ (or inverse of it). As a result, we proposed another independent optimization wherein the parameter represents information of $P_A$$+$$P_B$ herein. Owing to this novel approach to the BP, computational efficiency can be obtained by directly using physical objects. This idea of the physical implementation of TOW is applicable to various fields including constructing completely new analog computers~\cite{kimASTE}. 

\section*{Extended Prisoner's Dilemma Game}

Consider a situation wherein three people are arrested by police and are required to choose from the following five options:
\begin{itemize}
\item 
A: keep silent,
\item
B: confess (implicate him- or herself)
\item
C: implicate the next person  (circulative as 1,2,3,1,2,3,$\cdots$),
\item
D: implicate the third person (circulative as 1,2,3,1,2,3,$\cdots$),
\item
E: implicate both of the others.
\end{itemize}
According to the three person's choices, the `degree of charges' (from $0$ to $3$) are to be determined for every person. For example, the degree of charges are $(1,1,1)$ for the choice (person 1, person 2, person3) $=$ ($B$,$B$,$B$), $(1,1,0)$ for the choice ($A$,$B$,$C$), $(2,1,1)$ for the choice ($B$,$C$,$D$), $(2,2,1)$ for the choice ($C$,$D$,$D$), $(2,2,2)$ for the choice ($D$,$D$,$D$), $(3,1,1)$ for the choice ($B$,$D$,$D$) etc. For each pattern of degree of charges, a set of reward probabilities of each person are determined as follows:
\begin{itemize}
\item 
the $(0,0,0)$ is ($R2$,$R2$,$R2$),
\item 
the $(1,1,1)$ is  ($R1$,$R1$,$R1$), 
\item 
the $(2,1,1)$ or $(1,2,1)$ or $(1,1,2)$ is ($R$,$R$,$R$): the social maximum, 
\item 
the $(2,2,2)$ is ($P$,$P$,$P$): the Nash eqilibrium.
\end{itemize}
Otherwise, each difference between his or her degree and the minimum degree of the pattern determines a reward probability. If his/her degree is the same as the minimum degree, the reward probability is `T', otherwise `S'. Moreover, the difference between his or her degree and the minimum degree of the pattern is added to it. For example, the $(1,1,0)$ is ($S1$,$S1$,$T1$), the $(2,2,1)$ is ($S1$,$S1$,$T1$), the $(3,1,1)$ is ($S2$,$T2$,$T2$). Here, we set $T3$ $=$ $0.79$, $T2$ $=$ $0.76$, $T1$ $=$ $0.73$, $R$ $=$ $0.70$, $R1$ $=$ $0.60$, $R2$ $=$ $0.55$, $P$ $=$ $0.50$, $S1$ $=$ $0.40$, $S2$ $=$ $0.30$ and $S3$ $=$ $0.20$. Therefore, the social maximum is ($R$,$R$,$R$).
Here, it is assumed that police knows that there are a main suspect and two accomplices.
The complete list of reward probabilities is shown in Table~\ref{table:1}, \ref{table:2}, \ref{table:3} and \ref{table:4}.

\begin{table}[ht]
\caption{Reward probabilities in the Extended Prisoner's Dilemma Game.}
\label{table:1}
\begin{center}
\begin{tabular}{|c|c|c|} \hline \hline
  selection pattern       &  degree of charges  &  probability \\ \hline
( A, A, A ) &  ( 0, 0, 0 ) & 0.55 0.55 0.55 \\ \hline 
( A, A, B ) &  ( 0, 0, 1 ) & 0.73 0.73 0.40 \\ \hline 
( A, A, C ) &  ( 1, 0, 0 ) & 0.40 0.73 0.73 \\ \hline 
( A, A, D ) &  ( 0, 1, 0 ) & 0.73 0.40 0.73 \\ \hline 
( A, A, E ) &  ( 1, 1, 0 ) & 0.40 0.40 0.73 \\ \hline 
( A, B, A ) &  ( 0, 1, 0 ) & 0.73 0.40 0.73 \\ \hline 
( A, B, B ) &  ( 0, 1, 1 ) & 0.73 0.40 0.40 \\ \hline 
( A, B, C ) &  ( 1, 1, 0 ) & 0.40 0.40 0.73 \\ \hline 
( A, B, D ) &  ( 0, 2, 0 ) & 0.76 0.30 0.76 \\ \hline 
( A, B, E ) &  ( 1, 2, 0 ) & 0.73 0.30 0.76 \\ \hline 
( A, C, A ) &  ( 0, 0, 1 ) & 0.73 0.73 0.40 \\ \hline 
( A, C, B ) &  ( 0, 0, 2 ) & 0.76 0.76 0.30 \\ \hline 
( A, C, C ) &  ( 1, 0, 1 ) & 0.40 0.73 0.40 \\ \hline 
( A, C, D ) &  ( 0, 1, 1 ) & 0.73 0.40 0.40 \\ \hline 
( A, C, E ) &  ( 1, 1, 1 ) & 0.60 0.60 0.60 \\ \hline 
( A, D, A ) &  ( 1, 0, 0 ) & 0.40 0.73 0.73 \\ \hline 
( A, D, B ) &  ( 1, 0, 1 ) & 0.40 0.73 0.40 \\ \hline 
( A, D, C ) &  ( 2, 0, 0 ) & 0.30 0.76 0.76 \\ \hline 
( A, D, D ) &  ( 1, 1, 0 ) & 0.40 0.40 0.73 \\ \hline 
( A, D, E ) &  ( 2, 1, 0 ) & 0.30 0.73 0.76 \\ \hline 
( A, E, A ) &  ( 1, 0, 1 ) & 0.40 0.73 0.40 \\ \hline 
( A, E, B ) &  ( 1, 0, 2 ) & 0.73 0.76 0.30 \\ \hline 
( A, E, C ) &  ( 2, 0, 1 ) & 0.30 0.76 0.73 \\ \hline 
( A, E, D ) &  ( 1, 1, 1 ) & 0.60 0.60 0.60 \\ \hline 
( A, E, E ) &  ( 2, 1, 1 ) & 0.70 0.70 0.70 \\ \hline 
( B, A, A ) &  ( 1, 0, 0 ) & 0.40 0.73 0.73 \\ \hline 
( B, A, B ) &  ( 1, 0, 1 ) & 0.40 0.73 0.40 \\ \hline 
( B, A, C ) &  ( 2, 0, 0 ) & 0.30 0.76 0.76 \\ \hline 
( B, A, D ) &  ( 1, 1, 0 ) & 0.40 0.40 0.73 \\ \hline 
( B, A, E ) &  ( 2, 1, 0 ) & 0.30 0.73 0.76 \\ \hline 
( B, B, A ) &  ( 1, 1, 0 ) & 0.40 0.40 0.73 \\ \hline 
( B, B, B ) &  ( 1, 1, 1 ) & 0.60 0.60 0.60 \\ \hline 
( B, B, C ) &  ( 2, 1, 0 ) & 0.30 0.73 0.76 \\ \hline 
( B, B, D ) &  ( 1, 2, 0 ) & 0.73 0.30 0.76 \\ \hline 
( B, B, E ) &  ( 2, 2, 0 ) & 0.30 0.30 0.76 \\ \hline 
( B, C, A ) &  ( 1, 0, 1 ) & 0.40 0.73 0.40 \\ \hline 
\end{tabular}
\end{center}
\end{table}

\begin{table}[ht]
\caption{Reward probabilities in the Extended Prisoner's Dilemma Game.}
\label{table:2}
\begin{center}
\begin{tabular}{|c|c|c|} \hline \hline
  selection pattern       &  degree of charges  &  probability \\ \hline
( B, C, B ) &  ( 1, 0, 2 ) & 0.73 0.76 0.30 \\ \hline 
( B, C, C ) &  ( 2, 0, 1 ) & 0.30 0.76 0.73 \\ \hline 
( B, C, D ) &  ( 1, 1, 1 ) & 0.60 0.60 0.60 \\ \hline 
( B, C, E ) &  ( 2, 1, 1 ) & 0.70 0.70 0.70 \\ \hline 
( B, D, A ) &  ( 2, 0, 0 ) & 0.30 0.76 0.76 \\ \hline 
( B, D, B ) &  ( 2, 0, 1 ) & 0.30 0.76 0.73 \\ \hline 
( B, D, C ) &  ( 3, 0, 0 ) & 0.20 0.79 0.79 \\ \hline 
( B, D, D ) &  ( 2, 1, 0 ) & 0.30 0.73 0.76 \\ \hline 
( B, D, E ) &  ( 3, 1, 0 ) & 0.20 0.76 0.79 \\ \hline 
( B, E, A ) &  ( 2, 0, 1 ) & 0.30 0.76 0.73 \\ \hline 
( B, E, B ) &  ( 2, 0, 2 ) & 0.30 0.76 0.30 \\ \hline 
( B, E, C ) &  ( 3, 0, 1 ) & 0.20 0.79 0.76 \\ \hline 
( B, E, D ) &  ( 2, 1, 1 ) & 0.70 0.70 0.70 \\ \hline 
( B, E, E ) &  ( 3, 1, 1 ) & 0.30 0.76 0.76 \\ \hline 
( C, A, A ) &  ( 0, 1, 0 ) & 0.73 0.40 0.73 \\ \hline 
( C, A, B ) &  ( 0, 1, 1 ) & 0.73 0.40 0.40 \\ \hline 
( C, A, C ) &  ( 1, 1, 0 ) & 0.40 0.40 0.73 \\ \hline 
( C, A, D ) &  ( 0, 2, 0 ) & 0.76 0.30 0.76 \\ \hline 
( C, A, E ) &  ( 1, 2, 0 ) & 0.73 0.30 0.76 \\ \hline 
( C, B, A ) &  ( 0, 2, 0 ) & 0.76 0.30 0.76 \\ \hline 
( C, B, B ) &  ( 0, 2, 1 ) & 0.76 0.30 0.73 \\ \hline 
( C, B, C ) &  ( 1, 2, 0 ) & 0.73 0.30 0.76 \\ \hline 
( C, B, D ) &  ( 0, 3, 0 ) & 0.79 0.20 0.79 \\ \hline 
( C, B, E ) &  ( 1, 3, 0 ) & 0.76 0.20 0.79 \\ \hline 
( C, C, A ) &  ( 0, 1, 1 ) & 0.73 0.40 0.40 \\ \hline 
( C, C, B ) &  ( 0, 1, 2 ) & 0.76 0.73 0.30 \\ \hline 
( C, C, C ) &  ( 1, 1, 1 ) & 0.60 0.60 0.60 \\ \hline 
( C, C, D ) &  ( 0, 2, 1 ) & 0.76 0.30 0.73 \\ \hline 
( C, C, E ) &  ( 1, 2, 1 ) & 0.70 0.70 0.70 \\ \hline 
( C, D, A ) &  ( 1, 1, 0 ) & 0.40 0.40 0.73 \\ \hline 
( C, D, B ) &  ( 1, 1, 1 ) & 0.60 0.60 0.60 \\ \hline 
( C, D, C ) &  ( 2, 1, 0 ) & 0.30 0.73 0.76 \\ \hline 
( C, D, D ) &  ( 1, 2, 0 ) & 0.73 0.30 0.76 \\ \hline 
( C, D, E ) &  ( 2, 2, 0 ) & 0.30 0.30 0.76 \\ \hline 
( C, E, A ) &  ( 1, 1, 1 ) & 0.60 0.60 0.60 \\ \hline 
( C, E, B ) &  ( 1, 1, 2 ) & 0.70 0.70 0.70 \\ \hline 
\end{tabular}
\end{center}
\end{table}

\begin{table}[ht]
\caption{Reward probabilities in the Extended Prisoner's Dilemma Game.}
\label{table:3}
\begin{center}
\begin{tabular}{|c|c|c|} \hline \hline
  selection pattern       &  degree of charges  &  probability \\ \hline
( C, E, C ) &  ( 2, 1, 1 ) & 0.70 0.70 0.70 \\ \hline 
( C, E, D ) &  ( 1, 2, 1 ) & 0.70 0.70 0.70 \\ \hline 
( C, E, E ) &  ( 2, 2, 1 ) & 0.40 0.40 0.73 \\ \hline 
( D, A, A ) &  ( 0, 0, 1 ) & 0.73 0.73 0.40 \\ \hline 
( D, A, B ) &  ( 0, 0, 2 ) & 0.76 0.76 0.30 \\ \hline 
( D, A, C ) &  ( 1, 0, 1 ) & 0.40 0.73 0.40 \\ \hline 
( D, A, D ) &  ( 0, 1, 1 ) & 0.73 0.40 0.40 \\ \hline 
( D, A, E ) &  ( 1, 1, 1 ) & 0.60 0.60 0.60 \\ \hline 
( D, B, A ) &  ( 0, 1, 1 ) & 0.73 0.40 0.40 \\ \hline 
( D, B, B ) &  ( 0, 1, 2 ) & 0.76 0.73 0.30 \\ \hline 
( D, B, C ) &  ( 1, 1, 1 ) & 0.60 0.60 0.60 \\ \hline 
( D, B, D ) &  ( 0, 2, 1 ) & 0.76 0.30 0.73 \\ \hline 
( D, B, E ) &  ( 1, 2, 1 ) & 0.70 0.70 0.70 \\ \hline 
( D, C, A ) &  ( 0, 0, 2 ) & 0.76 0.76 0.30 \\ \hline 
( D, C, B ) &  ( 0, 0, 3 ) & 0.79 0.79 0.20 \\ \hline 
( D, C, C ) &  ( 1, 0, 2 ) & 0.73 0.76 0.30 \\ \hline 
( D, C, D ) &  ( 0, 1, 2 ) & 0.76 0.73 0.30 \\ \hline 
( D, C, E ) &  ( 1, 1, 2 ) & 0.70 0.70 0.70 \\ \hline 
( D, D, A ) &  ( 1, 0, 1 ) & 0.40 0.73 0.40 \\ \hline 
( D, D, B ) &  ( 1, 0, 2 ) & 0.73 0.76 0.30 \\ \hline 
( D, D, C ) &  ( 2, 0, 1 ) & 0.30 0.76 0.73 \\ \hline 
( D, D, D ) &  ( 1, 1, 1 ) & 0.60 0.60 0.60 \\ \hline 
( D, D, E ) &  ( 2, 1, 1 ) & 0.70 0.70 0.70 \\ \hline 
( D, E, A ) &  ( 1, 0, 2 ) & 0.73 0.76 0.30 \\ \hline 
( D, E, B ) &  ( 1, 0, 3 ) & 0.76 0.79 0.20 \\ \hline 
( D, E, C ) &  ( 2, 0, 2 ) & 0.30 0.76 0.30 \\ \hline 
( D, E, D ) &  ( 1, 1, 2 ) & 0.70 0.70 0.70 \\ \hline 
( D, E, E ) &  ( 2, 1, 2 ) & 0.40 0.73 0.40 \\ \hline 
( E, A, A ) &  ( 0, 1, 1 ) & 0.73 0.40 0.40 \\ \hline 
( E, A, B ) &  ( 0, 1, 2 ) & 0.76 0.73 0.30 \\ \hline 
( E, A, C ) &  ( 1, 1, 1 ) & 0.60 0.60 0.60 \\ \hline 
( E, A, D ) &  ( 0, 2, 1 ) & 0.76 0.30 0.73 \\ \hline 
( E, A, E ) &  ( 1, 2, 1 ) & 0.70 0.70 0.70 \\ \hline 
( E, B, A ) &  ( 0, 2, 1 ) & 0.76 0.30 0.73 \\ \hline 
( E, B, B ) &  ( 0, 2, 2 ) & 0.76 0.30 0.30 \\ \hline 
( E, B, C ) &  ( 1, 2, 1 ) & 0.70 0.70 0.70 \\ \hline 
\end{tabular}
\end{center}
\end{table}

\begin{table}[ht]
\caption{Reward probabilities in the Extended Prisoner's Dilemma Game.}
\label{table:4}
\begin{center}
\begin{tabular}{|c|c|c|} \hline \hline
  selection pattern       &  degree of charges  &  probability \\ \hline
( E, B, D ) &  ( 0, 3, 1 ) & 0.79 0.20 0.76 \\ \hline 
( E, B, E ) &  ( 1, 3, 1 ) & 0.76 0.30 0.76 \\ \hline 
( E, C, A ) &  ( 0, 1, 2 ) & 0.76 0.73 0.30 \\ \hline 
( E, C, B ) &  ( 0, 1, 3 ) & 0.79 0.76 0.20 \\ \hline 
( E, C, C ) &  ( 1, 1, 2 ) & 0.70 0.70 0.70 \\ \hline 
( E, C, D ) &  ( 0, 2, 2 ) & 0.76 0.30 0.30 \\ \hline 
( E, C, E ) &  ( 1, 2, 2 ) & 0.73 0.40 0.40 \\ \hline 
( E, D, A ) &  ( 1, 1, 1 ) & 0.60 0.60 0.60 \\ \hline 
( E, D, B ) &  ( 1, 1, 2 ) & 0.70 0.70 0.70 \\ \hline 
( E, D, C ) &  ( 2, 1, 1 ) & 0.70 0.70 0.70 \\ \hline 
( E, D, D ) &  ( 1, 2, 1 ) & 0.70 0.70 0.70 \\ \hline 
( E, D, E ) &  ( 2, 2, 1 ) & 0.40 0.40 0.73 \\ \hline 
( E, E, A ) &  ( 1, 1, 2 ) & 0.70 0.70 0.70 \\ \hline 
( E, E, B ) &  ( 1, 1, 3 ) & 0.76 0.76 0.30 \\ \hline 
( E, E, C ) &  ( 2, 1, 2 ) & 0.40 0.73 0.40 \\ \hline 
( E, E, D ) &  ( 1, 2, 2 ) & 0.73 0.40 0.40 \\ \hline 
( E, E, E ) &  ( 2, 2, 2 ) & 0.50 0.50 0.50 \\ \hline 
\end{tabular}
\end{center}
\end{table}

\end{document}